\documentclass[runningheads]{llncs}
\usepackage{graphicx}

\usepackage{tikz}
\usepackage{comment} 
\usepackage{amsmath,amssymb} 
\usepackage{color}
\usepackage{graphicx}
\usepackage{epsfig}
\usepackage{mathtools}
\usepackage{iac_pkg}
\usepackage{lipsum}
\usepackage{multirow}
\usepackage{tabularx, booktabs}
\usepackage{subfig}
\usepackage{color}
\usepackage{wrapfig}
\usepackage{xspace}

\definecolor{citecolor}{RGB}{34,139,34}
\usepackage[pagebackref=true,breaklinks=true,letterpaper=true,citecolor=citecolor,colorlinks,bookmarks=false]{hyperref}
\usepackage[capitalise]{cleveref}

\begin{document}
\pagestyle{headings}
\mainmatter
\def\ECCVSubNumber{--}  

\title{Two-branch Recurrent Network\\ for Isolating Deepfakes in Videos}

\titlerunning{Two-branch Recurrent Network for Deepfake Detection}
%
\author{Iacopo Masi\textsuperscript{*}\orcidID{0000-0003-0444-7646} \and
Aditya Killekar\textsuperscript{*} \and Royston Marian Mascarenhas \and Shenoy Pratik Gurudatt \and Wael AbdAlmageed
}

\authorrunning{I. Masi et al.}
%
\institute{USC Information Sciences Institute, Marina del Rey, CA, USA\\
\email{\{iacopo,killekar,royston,gurudatt,wamageed\}@isi.edu}\\
\href{https://www.youtube.com/watch?v=RspKj9DtM9U}{Demo of Our DeepFake Detection System}\\
\href{https://youtu.be/X3N8QjV15d8}{Video Presentation}}
\input{sections/00_teaser}
{\let\thefootnote\relax\footnotetext{\textsuperscript{*} indicates equal contribution}}
\maketitle
\begin{abstract}
The current spike of hyper-realistic faces artificially generated using \emph{deepfakes} calls for media forensics solutions that are tailored to video streams and work reliably with a low false alarm rate at the video level. 
We present a method for \emph{deepfake} detection based on a two-branch network structure that isolates digitally manipulated faces by learning to amplify artifacts while suppressing the high-level face content.
Unlike current methods that extract spatial frequencies as a preprocessing step, we propose a two-branch structure: one branch propagates the original information, while the other branch suppresses the face content yet amplifies multi-band frequencies using a \emph{Laplacian of Gaussian (LoG)} as a bottleneck layer.
To better isolate manipulated faces, we derive a novel cost function that, unlike regular classification, compresses the variability of natural faces and pushes away the unrealistic facial samples in the feature space.
Our two novel components show promising results on the \ffpp, Celeb-DF, and Facebook's DFDC preview benchmarks, when compared to prior work. We then offer a full, detailed ablation study of our network architecture and cost function. Finally, although the bar is still high to get very remarkable figures at a very low false alarm rate, our study shows that we can achieve good video-level performance when cross-testing in terms of video-level AUC.
\keywords{\deepfake detection, two-branch recurrent net, loss function}
\end{abstract}
\section{Introduction}\label{sec:intro}
\emph{Visual} misinformation has dramatically increased  on social networks and Internet~\cite{cnn_deepfakes}. Nonetheless, image manipulation is not new. Falsification of lithographs or photographs has been used for many years to reinforce political ideas or political characters~\cite{guera2018deepfake} or to practice censorship by erasing people from pictures. For instance, at the beginning of the twentieth century, political dissidents were assassinated and then erased from photographs through airbrushing during the Great Terror period in the Soviet Union~\cite{gellately2007lenin}.

In the modern era of digital pictures, perpetrators used commercial software and ``elbow grease'' to create realistic swapping of faces given a pair of still images. Although some of these results look very realistic, they involved a huge amount of manual work (on the order of hours) using a personal computer and an expensive raster graphics editor to produce just a single image~\cite{heller2018psBattles}. However, the effort required to produce face swaps diminished drastically during the last five years.  Democratized artificial intelligence (AI) made it very easy to produce highly realistic face swaps with a few clicks, giving the ability to non-experts to synthesize content with ``Hollywood-like'' effects just by simply using off-the-shelf applications~\cite{nirkin2018_faceswap}. The technology was quickly developed to process videos, transferring the identity of a subject from a \emph{source} video into a \emph{target} video. Unlike manual digital editing, face swapping in videos became effective and efficient, reaching hyper-realistic results, thanks to recent advances in data synthesis using Generative Adversarial Networks (GANs)~\cite{goodfellow2014generative}, Deep Convolutional Neural Networks (DCNN)~\cite{krizhevsky2012imagenet}, and AutoEncoders (AE)~\cite{kingma2013auto}. It also became easily available to non-experts through customized applications, such as DeepFaceLab~\cite{deepfacelab}, or even mobile applications, such as Zao~\cite{zao_app}. 

Face swapping has been superseded by \deepfakes  in which the original face is replaced with a victim's face with the intent of showing the victim to be saying something he/she never said. The fake video is usually very realistic so that the viewer believes that the swapped subject is the actual acting person in the video. Although in the beginning, \deepfakes were used to entertain users, they became popular to spread political chaos, revenge porn, and defamation. For these reasons, the rapid sharing of \deepfakes on the Internet became a threat to society leading to a common perception that \emph{seeing is no longer believing}~\cite{cnn_deepfakes}.

A recent report from DeepTrace~\cite{deeptrace} explains that the rate of increase in these fakes videos is 100\%  a year. Although \deepfakes initially appeared in 2017 on \texttt{reddit}, the report estimates that there are currently 14,678 realistic-looking yet fake videos, while the total number available in December 2018 was only 7,964. Given the current progress of AI and deep learning, the prediction is that this number may skyrocket in the near future.
In order to mitigate the proliferation of manipulated videos, we propose a deep learning architecture to detect hyper-realistic face manipulations. The paper makes the following contributions:

    $\diamond$ A two-branch representation extractor based on densely connected layers~\cite{huang2017densely} that learns to combine information from the color domain and the frequency domain using a multi-scale Laplacian of Gaussian (LoG) operator~\cite{burt1983laplacian}. The LoG operator suppresses the image content present in the low-level feature maps, acting as a band-pass filter to amplify artifacts.

    $\diamond$ A novel  loss function that encourages compactness of the representations of natural faces and pushes away manipulated faces for better, wider separation boundaries, which is different than recent methods that use binary cross-entropy for detecting face manipulations ~\cite{rossler2019faceforensics++v3,sabir2019recurrent}.
    
    $\diamond$ As a minor contribution, we argue that current metrics (accuracy) are improper for this problem, mainly for being very sensitive to class imbalance and failure to capture performance for web-scale applications. Therefore, we follow~\cite{korshunov2018deepfakes_df_timit,stehouwer2019detection} and report True Acceptance Rate ({TAR}) at low False Acceptance Rates ({FAR}). Also, besides standard area under receiver operating curve (AUC), we further propose global metrics at a low false alarm rate such as standardized partial AUC ({pAUC})~\cite{mcclish1989analyzing} and our truncated Area Under the Curve ({tAUC}).

We optimize our method for better generalization across datasets, reaching a good balance between bias and variance~\cite{pedro2000unified,valentini2004bias,domingos2012few}, i.e., performing remarkably on same dataset used for training~\cite{rossler2019faceforensics++v3} yet transferring reasonably  well across datasets~\cite{li2019celeb_v2,dolhansky_deepfake_2019}. Similar to only few works in literature~\cite{guera2018deepfake,sabir2019recurrent}, we also use sequential modeling for video-based detection. Our method processes sequences of aligned faces from a video, extracts discriminative features using the backbone, and performs recurrent modeling using bi-directional long short-term memory (LSTM) supervised by our new loss. The entire network is trained end-to-end so that the recurrent model back-propagates to the feature extractor. \cref{fig:teaser} shows the predictions of our system on face videos downloaded from the web when trained only on \ffpp. Our method is summarized in \cref{fig:pipeline}.

\section{Prior Work}\label{sec:related}
\minisection{Face forensics datasets and evaluation} Unlike the proliferation of face recognition datasets~\cite{LFWTech,klare2015pushing,kemelmacher2016megaface,msceleb_acmmm,guo2016ms}, there has been a lack of large-scale face forensics datasets in the community for both training and evaluation. Although face swapping can be cast as a splicing image forgery technique, and some generic forensics sets contain facial splicing and copy-move forgeries~\cite{heller2018psBattles}, earlier specific face manipulation detection tools~\cite{han2017two} have been mainly evaluated on still images.
Small-scale benchmarks released for deepfake detection were produced in controlled environments, e.g., DF-TIMIT ~\cite{korshunov2018deepfakes_df_timit,KorshunovICB2019} using $32$ subjects selected from the VidTIMIT~\cite{sanderson2009multi} database with the intent of studying the weaknesses of face detection and recognition technology, or UADFV~\cite{li2018ictu} that offers around $50$ bona fide and $50$ fake videos.

Only recently, Rossler \etal proposed several versions of \ffpp\cite{rossler2019faceforensics++v3}, a medium-scale collection of manipulated videos counting a total of $1.8$ million manipulated frames using four methods: FaceSwap, DeepFakes, Face2Face~\cite{thies2016face2face}, and NeuralTextures~\cite{thies2019neural}. The same dataset was augmented by Google Research with another set containing deepfake videos, i.e., Google Deepfake Detection (DFD)~\cite{google_dfd}. At the same time, Facebook and other firms joined efforts to create a competition to detect fakes on the web, releasing a preview dataset ``The Deepfake Detection Challenge (DFDC)''~\cite{dolhansky_deepfake_2019} along with new metrics for evaluation. With the exception of~\cite{Li_2019_CVPR_Workshops}, the interesting novel aspect is that performance is considered at a video-level instead of frame-level, effectively evaluating models at low false alarm rate. Before~\cite{dolhansky_deepfake_2019}, accuracy was the only metric used to measure fake detection performance with a few exceptions~\cite{korshunov2018deepfakes_df_timit,stehouwer2019detection}.
Despite these contributions, the perceived quality of the synthesized videos offered by these sets appears still lower compared to the videos circulating on the web, thus Li \etal recently released Celeb-DF~\cite{li2019celeb_v2} to produce hyper-realistic deepfakes, reporting the frame-level AUC as a metric. This benchmark is compelling, offering $5,369$ high quality videos for a total of $2.1$M frames. 

\minisection{Detection of face manipulations} Although image forensics has been widely studied for a long time~\cite{farid2016photo}, \deepfakes is recent technology and thus several orthogonal works have been proposed lately for solving the problem of detecting face manipulations. Methods for deepfake detection can be roughly categorized in two macroscopic groups --- (i)  discriminative classifiers that use diverse semantic inconsistencies of the head and face; and (ii) data-driven approaches directly learning a discriminative function from data.
Considering the first group, Agarwal \etal built person-specific classifiers~\cite{Agarwal_2019_CVPR_Workshops} using one-class support vector machines (SVM) and features computed from Action Units (AU) and 3D head pose movements. Similarly, Li \etal~\cite{li2018ictu} used the observation that initial versions of \deepfakes were not blinking.  Later they extended the work to check for the inconsistency of 3D head poses. They also trained a DCNN though they used as negative samples faces undergoing warping artifacts to simulate the deepfake stitching process~\cite{Li_2019_CVPR_Workshops}, while \cite{matern2019exploiting} used hand-crafted visual features to amplify artifacts. Regarding the second group, XceptionNet~\cite{chollet_xception:_2017} has been widely used~\cite{rossler2019faceforensics++v3}, while ~\cite{afchar2018mesonet} used a variant of Inception module to capture both micro and mesoscopic features. Han \etal ~\cite{han2017two} used a two-branch structure similar to ours, yet, unlike our method, performed late fusion between a RGB branch and steganalysis features, using triplet loss for supervision. Later methods employed multi-task learning~\cite{nguyen2019multi} with an encoder-decoder similar to~\cite{cozzolino2018forensictransfer} and capsule networks~\cite{nguyen2019capsule}. For a complete survey, we refer to~\cite{li2019celeb_v2,nguyen2019deep} and the recent work in~\cite{tolosana2020deepfakes,verdoliva2020media}.

\minisection{GAN synthesis detection} Finally, a parallel line of related research~\cite{zhang2019detecting,yu2019attributing,cozzolino2018forensictransfer} \cite{verdoliva2019extracting,marra2019gans} is detecting entirely GAN-synthesized face images, e.g., using StyleGAN~\cite{karras2019style}.  Our work shares similar traits with the very recent research by Yu \etal ~\cite{yu2019attributing} with some major differences. We focus on detecting \deepfakes, while \cite{yu2019attributing}'s interest is in modeling GAN fingerprints. More importantly, our method is composed of a two-branch structure that fuses RGB information with the frequency domain.
\section{Method}\label{sec:method}
The objective is to learn a classifier for the detection of manipulated faces, squishing a set of aligned video frames\footnote{Throughout this paper $\bI$ indicates a sequence (or window) of aligned faces from video frames of cardinality $F$.} $\bI \in \mathbb{R}^{H \PLH W\PLH 3\PLH F}$ to an embedding $\net(\bI) \in \mathbb{R}^{D}$ so that the representations of natural faces are compact around a reference centroid $\bc$ and manipulated faces are spread out, ensuring a large margin between tampered and untamperd faces. In \cref{sec:arch} we introduce a two-branch backbone representation extractor $\net(\cdot)$ based on densely connected layers~\cite{huang2017densely}. $\net$ learns to fuse different representations obtained using regular convolutional filters $\net_{\text{RGB}}$ and representations extracted using multi-scale Laplacian of Gaussian~\cite{burt1983laplacian} kernels $\net_{\text{LoG}}$ (\cref{sec:deep_log}).
The combined features maps are then fed to the backbone that ends with a bi-directional Long Short-Term Memory (LSTM) for temporal modeling. $\net(\bI)$ indicates the concatenated output from the two bidirectional LSTM streams. The entire recurrent model is supervised through a novel formulation. Unlike recent methods~\cite{rossler2019faceforensics++v3,sabir2019recurrent} that use classification losses for detection, in \cref{sec:loss} we introduce a loss function that encourages the compactness of the representations of untampered faces, while distancing the representations of manipulated faces, for wider separation boundaries. At test-time, given an input sequence $\bI$, the method obtains the distance $\norm{\net(\bI)-\bc}_2$; the larger the distance the higher the likelihood of the sample being manipulated.

\begin{figure*}[tb]
    \begin{center}
    \includegraphics[width=\linewidth]{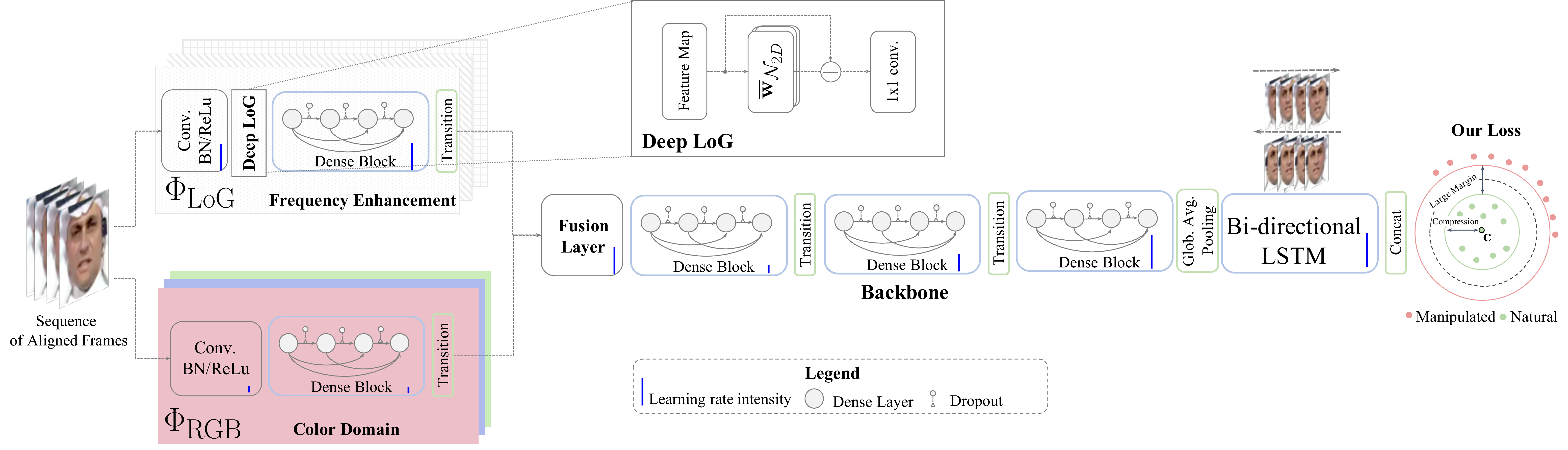}
    \end{center}  
    \caption{{\bf Our video-based face manipulation detection architecture.} A face sequence is processed by two independent DenseBlocks: one subject to a Deep Laplacian of Gaussian (Deep LoG) layer (frequency enhancement) and the second is a branch that works in the color domain. The two feature maps are fused so that a backbone of dense blocks learns a rich representation. The architecture uses dropout after each DenseLayer and a different learning rate per layer to mitigate overfitting. Our architecture ends with a bi-directional LSTM layer supervised using a novel loss formulation.
    }
\label{fig:pipeline}
\end{figure*}

\subsection{Network Architecture and Optimization}\label{sec:arch}

\minisection{Architecture} The basic network architecture $\Phi$ is derived by the recent work in~\cite{sabir2019recurrent} with major modification. The network takes as input RGB faces but consists of two different branches: a regular DenseBlock ~\cite{huang2017densely} that learns to process color domain data $\net_{\text{RGB}}$ and another parallel DenseBlock $\net_{\text{LoG}}$ with unshared weights that learns to discard visual face content by applying a Laplacian of Gaussian (LoG) filter to the low-level feature maps. The two feature maps are aligned and have a resolution of $28\PLH28$ with 128 planes. These maps are fused together with point-wise convolution with two groups~\cite{ioannou2017deep} such that each group of convolutional filters independently refines and fuses the information for the three downstream DenseBlocks. All the DenseBlocks end with a Transitional layer except for the one prior the LSTM. We discard the final linear classification layer and reduce the final feature map to a feature vector with dimension $1024$ using global average pooling. Dropout with a probability of $0.2$ is applied at the end of each DenseLayer to avoid overfitting.

\minisection{Optimization} Instead of optimizing the entire network with a single learning rate, which may overfit given the large parameter space of DenseNet, we employ a strategy that sets different learning rates per DenseBlock. In particular, given the global learning rate $\mu$, we define a decay for the DenseBlocks so that downstream layers incorporate gradients quickly  while upstream layers change less drastically. The learning rate decay is defined as $\mu_L\doteq\mu\cdot 1/(2^L)$ for the entire network with the exception of the DenseBlock of the Deep LoG and the layer in charge of fusing the two-branches. The parameters of fusion layers are updated faster, with the global rate $\mu$, since that they have to be adapted to the  frequency domain information. The blue bars in \cref{fig:pipeline} indicate the intensity of learning rate for each layer. We initialize all the layers from pre-trained weights from ImageNet except those of the LSTM, which are instead initialized from a uniform distribution.

\minisection{Video-based stratified sampling} Since inter-video variations are stronger than intra-video, we define a ``stratified epoch'' as the set of sequences obtained by sampling a  \emph{random sequence once from all the videos} in the training set. Stratified training ensures that each mini-batch includes a diversified set of training samples, avoiding including similar sequences from the same video within the mini-batch if we use simple random sampling. 
\begin{figure*}[tb]
\centering
    \includegraphics[trim={0 0 0 0},clip,width=.85\linewidth]{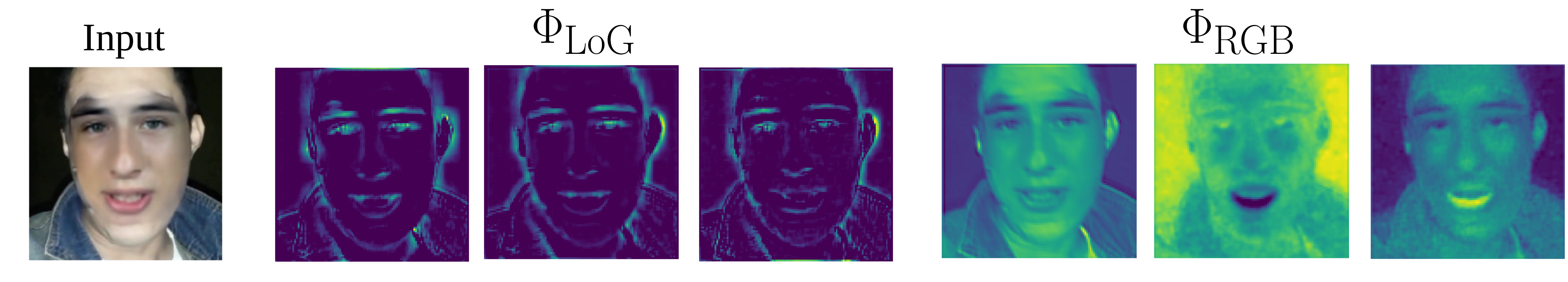}
    \caption{
    {\bf Diverse feature maps.} A diverse set of feature maps is obtained in the two distinct branches. The three representative feature maps after the first convolutional layer and our Deep LoG layer from $\net_{\text{LoG}}$ are shown on the left with the input. The feature maps on the right show the response from $\net_{\text{RGB}}$.
    }
   \label{fig:feat_maps}
\end{figure*}

\subsection{Deep Laplacian of Gaussian}\label{sec:deep_log}
\minisection{Motivation} Zhang \etal~\cite{zhang2019detecting} show that image manipulations leave medium- and high-frequency traces, and since most of the \deepfakes methods reconstruct a face with an average pixel-wise $\ell_2$ loss (Mean Square Error (MSE)), producing somewhat blurry (low-frequency) facial features, and because of the up-sampling commonly used in the decoder part. According to these observations, we design a custom layer to propagate multi-band frequency information inside the network with the goal of suppressing high-level face content, thereby amplify artifacts.

\minisection{Implementation} Without loss of generality, given a feature map $\bx$, we apply a Laplacian of Gaussian (LoG)~\cite{burt1983laplacian} as follows. Input tensor $~\mbf{x}$ (which can be in the first layer the input image alone or directly higher feature maps) is processed by two sets of convolutional filters: {fixed, non-learnable} filters $\overline{\bw}_{\mathcal{N}_{2D}}$, as a 2D Gaussian kernel and a dimensionality reduction filter $\bw_{1\PLH1}$ that maps back the dimensionality to the input expected by the next layer. The layer shares a similar design with~\cite{xie2019feature} although in our case (1) the objective is to suppress global information from the face, not to suppress noise from adversarial samples and  (2) the skip-connection is used to \emph{remove} information while~\cite{xie2019feature} used it to ease the training.
As described in \cref{eq:deep_log}, the output feature map $\bx^\prime$  is then obtained as shown in \cref{eq:deep_log}
\begin{equation}
\setlength\abovedisplayskip{5pt}
\setlength\belowdisplayskip{5pt}
\bx^\prime = \bw_{1\PLH1}\Big(\bx - \text{up}\big(\text{down}(\overline{\bw}_{\mathcal{N}_{2D}}~\bx)\big)\Big),
\label{eq:deep_log}
\end{equation}
where $\text{up}(\cdot), \text{down}(\cdot)$ indicate upsampling and downsampling, respectively, of the tensor across multiple scales $S$$=$$3$.
Assuming that the LoG is inserted into a layer with $K$ input planes and $K^{\prime}$ output planes, then internally the tensor depth dimension becomes  $K\xrightarrow{\text{LoG}}S~K\xrightarrow{\bw_{1\PLH1}}K^{\prime}$.
\cref{fig:feat_maps} shows the difference in the feature maps between the new LoG branch and the regular RGB branch after the first convolutional layer. All the feature maps are taken from the same filters, so they are aligned across branches. Each map is normalized $\in [0,1]$. We can see that most of the energy in the response for the $\net_{\text{LoG}}$ case is around the edges, the $\net_{\text{RGB}}$ counterpart instead focuses more on the global structure of the face.

\subsection{Loss Function to Isolate Manipulated Faces}\label{sec:loss}
\minisection{Motivation} The majority of previous work on face manipulation detection \hfill~\cite{zhang2019detecting,rossler2019faceforensics++v3,sabir2019recurrent} uses standard cost functions adopted from classification. Cozzolino \etal~\cite{cozzolino2018forensictransfer} recently made an effort toward representation disentanglement and generalization for face manipulation detection. Unlike previous work summarized in \cref{sec:related}, we propose a new loss for better isolating manipulated faces inspired by recent work on one-class classifiers, such as one-class Deep Support Vector Data Description (Deep SVDD)~\cite{ruff2018deep}. The new formulation induces compactness of the embedding space for sequences of unmanipulated faces. However, unlike~\cite{ruff2018deep}, the proposed loss employs manipulations synthesized by a few generators as negative samples enforcing a larger margin to the natural face sequences.

\begin{figure*}[tb]
\centering
\subfloat[Loss idea]{
    \includegraphics[trim={0 0 0 0},clip,width=0.2\linewidth]{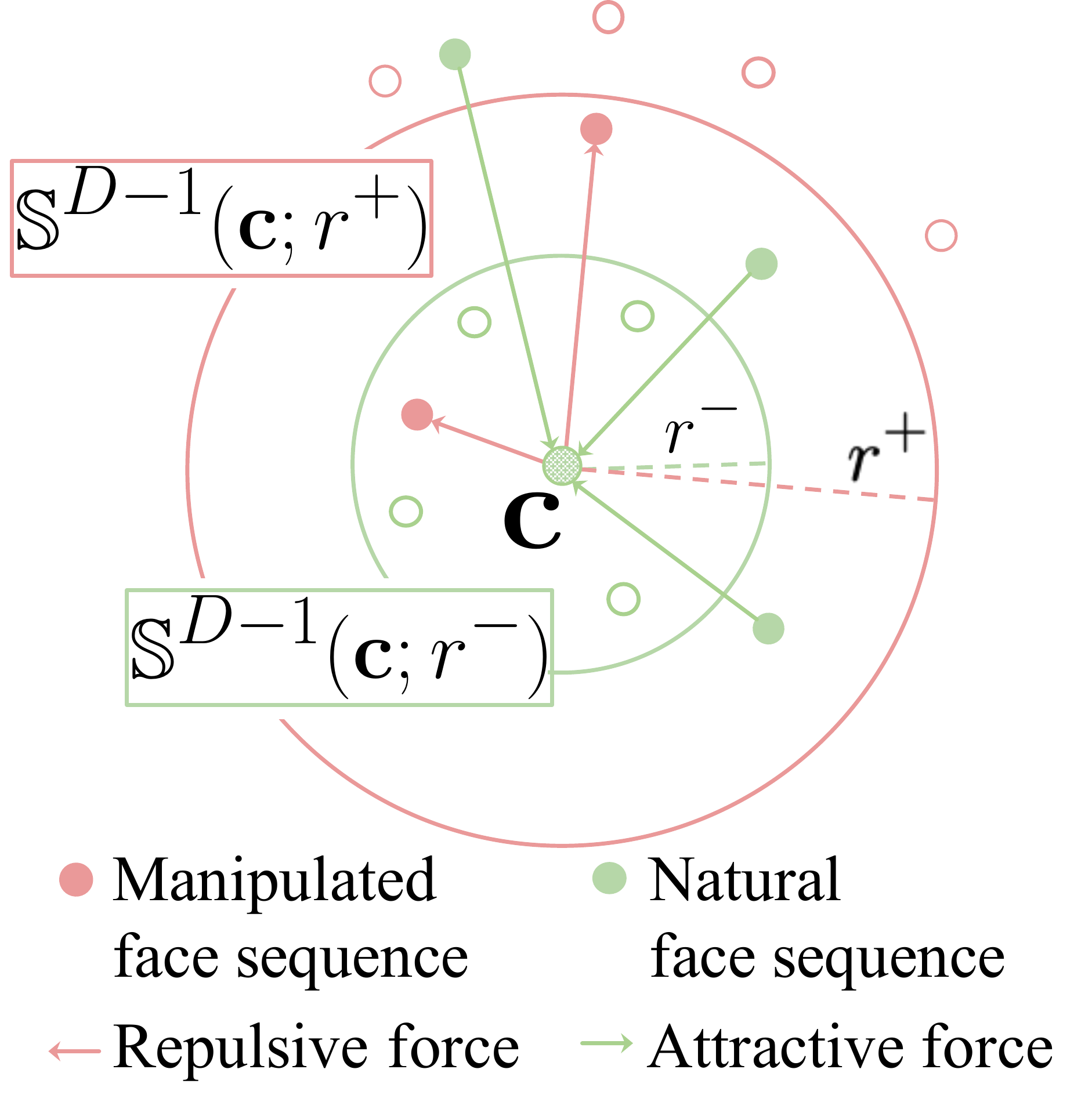}
    \label{fig:loss_idea}
}
\subfloat[Feature space]{
    \includegraphics[trim={0 0 0 0},clip,width=0.21\linewidth]{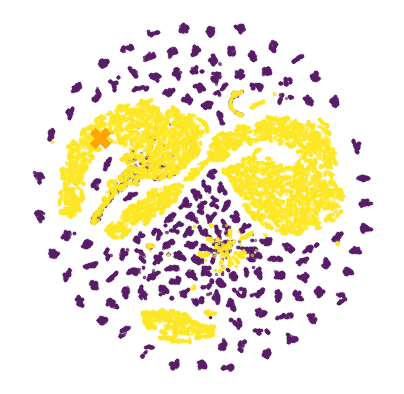}
    \label{fig:loss_train}
} \quad
\subfloat[Logit distribution]{
    \includegraphics[trim={0 0 0 0},clip,width=0.22\linewidth]{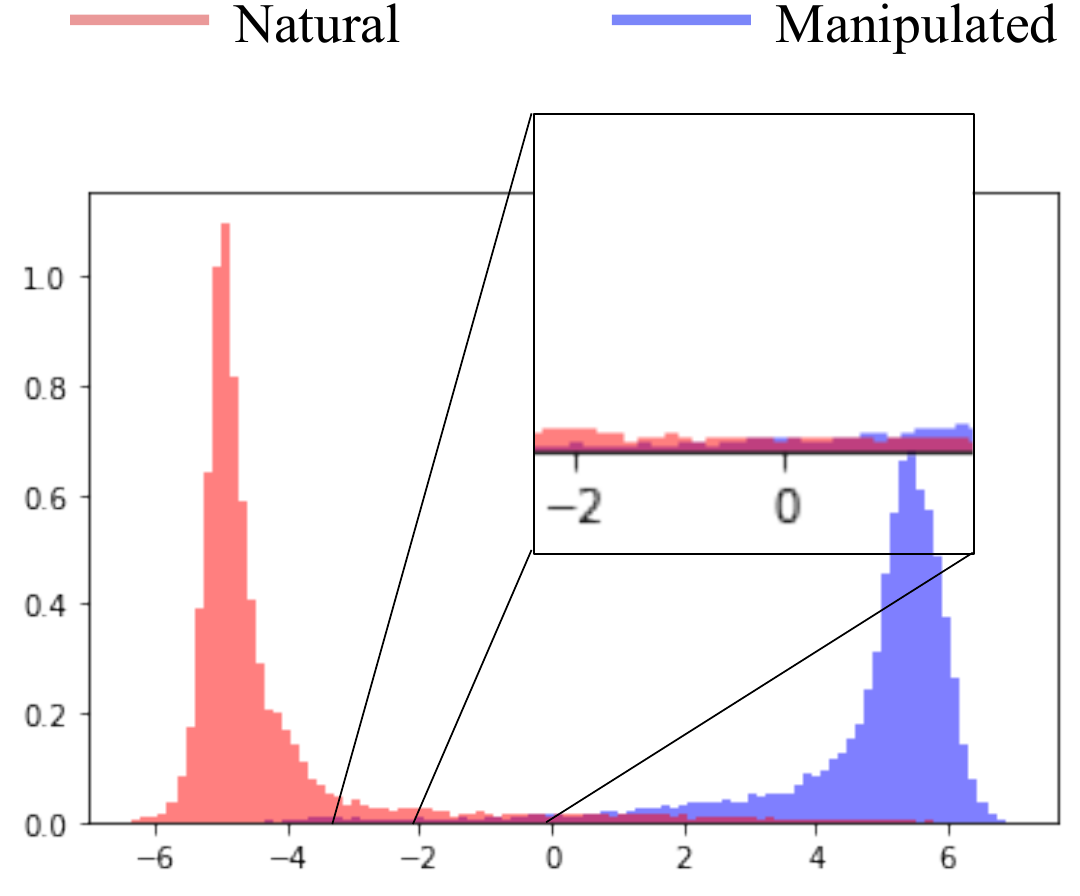}
    \label{fig:loss_logit}
}
\subfloat[Our distribution]{
    \includegraphics[trim={0 0 0 0},clip,width=0.22\linewidth]{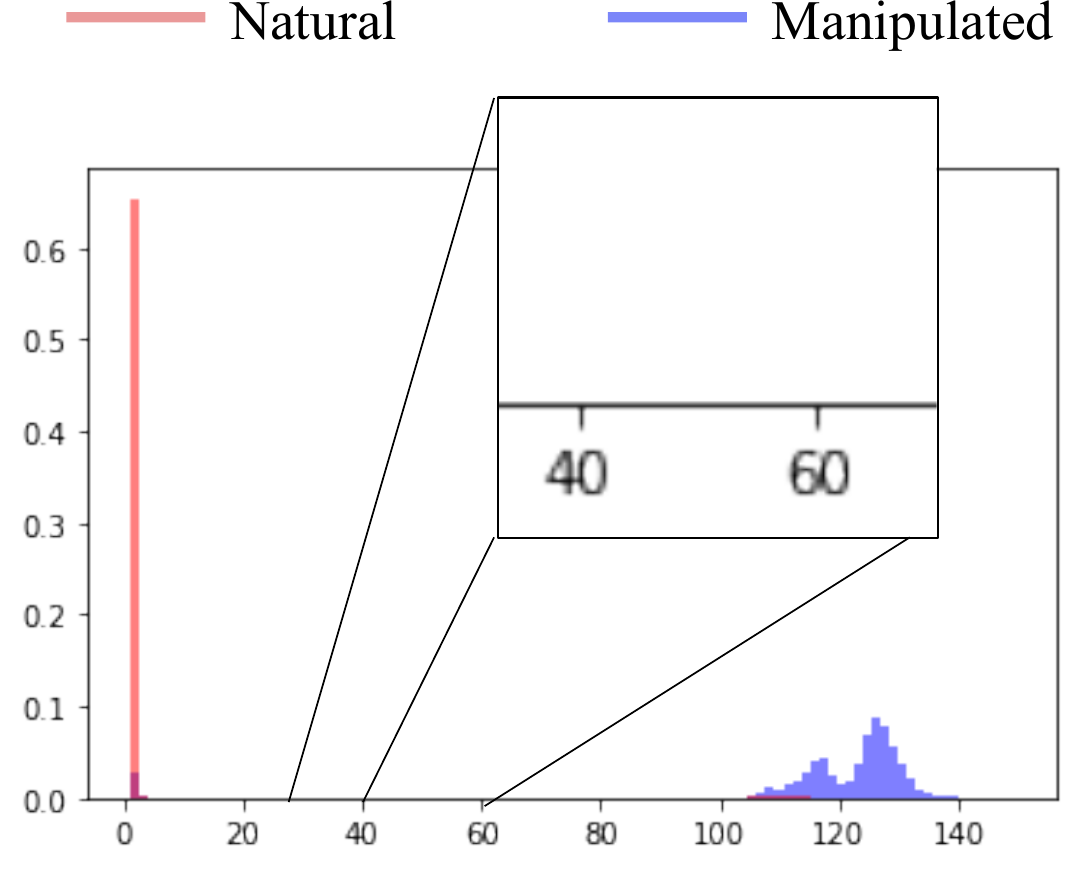}
    \label{fig:loss_distance}
}
\caption{
{\bf Loss formulation.} (a) The loss induces compression of the natural face sequences within a inner hypersphere placing easier samples close to $\mbf{c}$ and tougher samples at the boundary; meanwhile it induces a large margin forcing the manipulated face sequences outside the outer hypersphere (b) t-SNE~\cite{maaten2008visualizing} visualization of the feature space on the test set between natural faces (yellow) and \deepfakes (violet). The center $\bc$ is shown as an orange cross. (c) Genuine-Impostor distribution of logits with binary cross-entropy and (d) with our loss function: imposing a wider margin induces less confusion in the distribution.
}
\label{fig:loss}
\end{figure*}

\minisection{Formulation} More formally, we optimize the entire recurrent network defined in \cref{sec:arch} through a cost function that organizes the feature space such that the variability of sequences of natural faces is compacted toward a reference center while the representations of manipulated face sequences are placed far apart at the boundaries of the feature space.
Before training, we begin by pre-computing a reference center $\bc \in \mathbb{R}^D$ by averaging the encodings of all the natural, unmanipulated face sequences in the training set. The encodings are obtained by taking the responses of our entire architecture with two-branches and the bi-directional LSTM before training. The concatenated bidirectional LSTM features are extracted using the same network that is pre-trained. The two-branches and the backbone are pre-trained on ImageNet. When the training starts, all the features are aligned to this predefined embedding space. 
Then we define two hyperspheres centered around $\bc$ to constrain the feature space so that natural faces lie within $\mathbb{S}^{D-1}(\bc;r^-)$, while manipulated faces are kept outside $\mathbb{S}^{D-1}(\bc;r^+)$.
The loss induces compression on the regular faces embeddings. However, unlike ~\cite{ruff2018deep}, we avoid reducing all samples to a single high-dimensional point and mitigate overfitting by requiring compression up to an internal inner margin defined by the radius $r^-$ of the first hypersphere. Furthermore, the proposed loss enforces sequences of manipulated faces to be kept outside the second hypersphere defined by the radius $r^{+}$.
The loss $\mathcal{L}$ given a mini-batch $\Omega \in \mathbb{R}^{H \PLH W\PLH 3\PLH F\PLH B}$ of face sequences is defined as shown in \cref{eq:loss}:
\setlength\abovedisplayskip{5pt}
\setlength\belowdisplayskip{5pt}
\begin{multline}
\mathcal{L} =
    \frac{1}{|\Omega_{\text{nat.}}|} \sum_{i\in \Omega_{\text{nat.}}} \max\Big(0,\norm{\net(\bI_i)-\mbf{c}}_2-r^-\Big) + \\
    + \frac{1}{|\Omega_{\text{man.}}|} \sum_{j\in \Omega_{\text{man.}}} \max\Big(0,r^+ - \norm{\net(\bI_j)-\mbf{c}}_2\Big),
\label{eq:loss}
\end{multline}
where $\Omega_{\text{nat.}, \text{man.}}$ selects natural and manipulated face samples, respectively. For this loss to be valid, it has to hold that $0<r^-<r^+$ and the margin imposed between the two classes is $m=r^+-r^-$. The values of the two radii have to be set according to the dimensionality $D$ of the feature embedding. The loss mitigates the problem of class imbalances by normalizing each term by its cardinality.
Further, the second margin $r^{+}$ is essential to the loss because the network may chose to lower the cost just by pushing the negative samples indefinitely, without inducing compression on the natural faces.

\cref{fig:loss_idea} illustrates the basic idea of the proposed loss, and \cref{fig:loss_train} demonstrates the feature space of the test set of natural faces vs \deepfakes. The features are mapped to $\mathbb{R}^2$ using t-SNE~\cite{maaten2008visualizing} optimizing a plain DenseNet model. Natural faces are compressed while manipulated faces lie at the boundaries. The clusters formed by videos are visible for the manipulated faces.
\cref{fig:loss_logit} shows the genuine and impostor distribution of the logits at inference time for a model trained for discerning real faces from \deepfakes using binary cross-entropy on \ffpp~\cite{rossler2019faceforensics++v3}. Although the distribution presents two peaks corresponding to real and \deepfakes faces, the variance of those distribution is not minimized, and, more importantly, real face logits are spread out toward the manipulated faces thereby negatively affecting the detection rate at a low false alarm regime. In contrast, \cref{fig:loss_distance} offers the distribution of the distances from the center $\bc$ for the two classes. Using the proposed loss we achieved compression of the natural faces and a clear separation from the manipulated faces, visible when zooming in a highly confusing region.

\minisection{Interpretation} \cref{eq:loss} shares similar traits with the formulation in~\cite{ruff2018deep} with a few key differences.  First, we have a secondary term for supervision for abnormal cases. Second,  we have margins that avoid overfitting and better separate the two classes. The loss function also resembles the classic formulations found in deep metric learning such as contrastive loss functions~\cite{Wu_2017_ICCV}, although in our case the optimization is better constrained since the network is allowed to ``move'' only $\net(\bI)$ while $\bc$ is kept fixed. Finally, we spare the sampling of pairs or even triplets~\cite{schroff2015facenet}  which significantly reduces training complexity. Our loss differs from recent formulations: ~\cite{wen2016discriminative} uses softmax while we do not; it also sets one center for each class while we have a single center for both classes; finally, unlike us, ~\cite{wen2016discriminative} updates the centers while training. The work in~\cite{wang2018additive} enforces angular margin whereas ours uses radial distance margin;~\cite{wang2018additive} induces compactness in all the classes while ours only on natural face sequences.
\section{Experimental Evaluation}\label{sec:expts}
\minisection{Benchmarks and metrics} Ablation studies and comparisons are conducted on (1) \ffpp~\cite{rossler2019faceforensics++v3}, (2) Celeb-DF~\cite{li2019celeb_v2}, (3) and the Deepfake Detection Challenge (DFDC) Preview Dataset~\cite{dolhansky_deepfake_2019}. We report results at the video-level and also at the frame-level. Given that our method works at a sequence level, when comparing to other methods, we made sure that the number of samples prior computing the ROC is the same for all methods when comparing at the frame-level or, at least, that that all methods observed the same quantity of data.
Further, we use standard metrics such as True Acceptance Rate ({TAR}) at low False Acceptance Rates ({FAR}), similar to~\cite{korshunov2018deepfakes_df_timit,stehouwer2019detection}. Besides standard area under receiver operating curve (AUC), we further use global metrics yet at a low false alarm rate such. These metrics can shed light on performance in realistic operational  scenarios, thereby requiring detectors to operate at a very low false alarm rate and raising the bar for the community. We used the standardized partial AUC or {pAUC}~\cite{mcclish1989analyzing} and our {tAUC}, that is defined as AUC yet taking into consideration only the low false alarm rate up to a cut-off point $\text{FAR}_{\tau}$, thereby ignoring high false alarm rates. {tAUC} is computed as the ratio between the area of TARs up to a given low $\text{FAR}_{\tau}$ normalized by the total area up to the $\text{FAR}_{\tau}$ value. Given $\mathcal{F}_{\tau}=\{0,\ldots,\text{FAR}_{\tau}\}$, then tAUC at an operating point $\tau$ is defined as $
\text{tAUC}_{\tau} \doteq \frac{\sum_{i \in \mathcal{F}_{\tau} } \text{TAR}_i}{\lvert\mathcal{F}_{\tau}\rvert}.$ 

\minisection{Implementation and Hyper-Parameters} Unless otherwise stated, we used the following settings. The global learning rate $\mu$ is $\on{1e-03}$ using the Adam optimizer and the results are produced with LSTM. The learning rate is decreased three times by a factor of $10$. We decrease it every time the validation loss does not decrease after $50$ stratified epochs. We used a weight decay of $\on{1e-06}$. The final global average pooling flattening the spatial dimension gives a descriptor with dimensionality ${1024}$ transformed into $D$=$128$ by the LSTM. The final dimensionality considered in the loss is $2D$\footnote{The dimensionality is doubled since the results of the bi-directional streams are concatenated.} and the two radii $r^{\{-,+\}}\doteq\{\on{0.042},\on{1.638}\}$ have to be optimized together and cross-validated on a validation set. In high-dimensional space, the volume of the hyper-sphere decreases when the feature descriptor dimension $D$ increases~\cite{weisstein2002hypersphere}: thus, if $D$ does change, the radii have to be changed accordingly. By increasing the dimensionality $D$ of the final feature, the radii have to be increased as well to compensate for the diminished hyper-volume of the hyper-sphere.
The cardinality $F$ of the sequence of aligned frames as input to the recurrent model is 10. Since the sequential modeling is trained on sampled FF\Plus\Plus~data, at inference time we take 1 frame over 7 to build the sequence.
Faces are aligned with dlib~\cite{king2009dlib}. If alignment fails, we revert back to~\cite{bulat2017far}. In case of multiple detected faces, we select the largest detected face.
Since \ffpp has imbalanced labels (1:4), we oversample the natural faces twice and undersample randomly faces for each manipulation with a factor of two to get a proper balance, when training with multiple manipulations. We used average to perform video-level evaluation to aggregate all the scores within a video for all methods. When doing cross-testing, we use always the same model trained on FF\Plus\Plus~on the four manipulations on high compression (c40).

\subsection{\ffpp (FF\Plus\Plus)}\label{sec:ffpp}
\minisection{Settings} When training and evaluating on FF\Plus\Plus, we follow the sampling strategy mentioned in~\cite{rossler2019faceforensics++v3} that selects 270 frames/video for the training and 110 frames/video for validation and testing. We evaluated both medium compression (c23) and high compression levels (c40) subsets.

{
\begin{table}[tb]
\centering
\subfloat[]{
    \resizebox{0.31\linewidth}{!}{
      \begin{tabular}{lcccc}
        \toprule
        & \multicolumn{3}{c}{{\normalsize Deepfakes \texttt{c40} - wo/ LSTM}}\\
        \cmidrule(l){2-4} 
         ~ & tAUC$_{1\%}$ & tAUC$_{10\%}$ & TAR$_{1\%}$ \\
         \cmidrule(l){1-1} 
        Encoder~\cite{sabir2019recurrent}
        & 57.63 & 86.61 & 81.50 \\ 
        \cmidrule(l){1-1}
        \emph{+ft,+drop,+loss} 
        & {76.04} & {89.91} & {92.84} \\ 
      \bottomrule
      \end{tabular}
    }
      \label{table:ablation_ffpp_c40}
}
\subfloat[]{
    \resizebox{0.31\linewidth}{!}{
      \begin{tabular}{lccc}
        \toprule
        & \multicolumn{3}{c}{{\normalsize Deepfakes \texttt{c23} - wo/ LSTM}}\\
        \cmidrule(l){2-4}
          & tAUC$_{1\%}$ & tAUC$_{10\%}$ & TAR$_{1\%}$ \\  
        Single-Branch & 56.78 & 61.14   & 99.34\\  
        \rule{0pt}{2ex}
        Two-Branch    &  61.70 & 70.80  & 98.34 \\
      \bottomrule
      \end{tabular}
    }
      \label{table:ablation_ffpp_c23}
}
\subfloat[]{
    \resizebox{0.31\linewidth}{!}{
      \begin{tabular}{c@{~}c@{~}c@{~}c@{~}c@{~}c@{~}}
        \toprule
        & \multicolumn{5}{c}{{\normalsize Deepfakes \texttt{c40} - w/ LSTM}}\\
        \cmidrule(l){3-6} 
       LSTM & LSTM & \( \net{\text{{\tiny RGB}}} \)         & f.t. &  tAUC$_{1\%}$ &  TAR$_{1\%}$ \\
        hid.~nodes  & fusion      & \(  \circ ~\net{\text{{\tiny LoG}}} \) & +dropout &  &  \\
         128 & \texttt{cat} & \texttt{conv1x1$_{g=1}$} & --- &  57.21  & 83.33 \\
         128 & \texttt{cat} & \texttt{conv1x1$_{g=1}$} & \checkmark & 73.58  & 87.38 \\
         128 & \texttt{sum} & \texttt{conv1x1$_{g=2}$} & \checkmark & 67.49 & 83.81 \\
         256 & \texttt{cat} & \texttt{conv1x1$_{g=2}$} & \checkmark & 76.35 & 87.14 \\
         128 & \texttt{cat} & \texttt{conv1x1$_{g=2}$} & \checkmark & 81.53  & 92.54 \\
      \bottomrule
      \end{tabular}
    }
      \label{table:ablation_ffpp_c40_lstm}
}
  \caption{{\bf Ablation study on FF\Plus\Plus.} (a) Testing metrics obtained by training our model under different settings on the FF\Plus\Plus~\cite{rossler2019faceforensics++v3} under the highest compression level \texttt{c40} without LSTM, ablating our optimization and the loss function. (b) Ablation experiments showing the impact of the two branches under the medium compression level \texttt{c23}. (c) Ablation experiments using LSTM on \texttt{c40}.
  }
\end{table}
}
\begin{table}[tb]
\centering
	\resizebox{\linewidth}{!}{
		\begin{tabular}{lcccc@{~~~~}cccc@{~~~~}cccc@{~~~~}cccc}\toprule
			&  \multicolumn{8}{c}{HQ (\texttt{c23})}  & \multicolumn{8}{c}{LQ (\texttt{c40})}\\
			&  \multicolumn{4}{l}{Frame}  & \multicolumn{4}{l}{Video}   &  \multicolumn{4}{l}{Frame}  & \multicolumn{4}{l}{Video}  \\
			&  \multicolumn{4}{l}{Level ($\sim$70K samples)}  & \multicolumn{4}{l}{Level (700 samples)}   &  \multicolumn{4}{l}{Level ($\sim$70K samples)}  & \multicolumn{4}{l}{Level (700 samples)}  \\
			Methods & AUC & pAUC$_{10\%}$ & tAUC$_{10\%}$ & TAR$_{10\%}$ & AUC & pAUC$_{10\%}$ & tAUC$_{10\%}$ & TAR$_{10\%}$ & AUC & pAUC$_{10\%}$ & tAUC$_{10\%}$ & TAR$_{10\%}$ & AUC & pAUC$_{10\%}$ & tAUC$_{10\%}$ & TAR$_{10\%}$  \\
 			\cmidrule(l){1-1}
 			DSP-FWA \cite{Li_2019_CVPR_Workshops}  & 56.89 & 51.33 & 7.47 & 14.60 & 57.49 & 51.59 &7.48 & 15.00 
 			& 59.15 & 52.04 & 8.82 & 17.30 & 62.34& 51.93 & 9.82 &  22.14  \\
 			Xception~\cite{rossler2019faceforensics++v3}   & 92.30 & 87.71 & \tbf{73.34} & 81.21 & 
 		    92.50 & 89.20 & 58.21 & 82.85 & 83.93 & \tbf{74.78} & \tbf{45.92} & \tbf{63.25} & 86.75  & \tbf{79.10}  & 39.06  &   68.75\\
 			Ours                                              & \tbf{98.70} & \tbf{97.43} & 65.29 & \tbf{97.95} &
 			\tbf{99.12} & \tbf{98.41} & \tbf{86.10} & \tbf{98.21} & \tbf{86.59} & 69.71 & 40.41        & 62.48       & \tbf{91.10} & 76.57  & \tbf{51.18} & \tbf{72.85}\\
			\cmidrule(l){1-17}
		\end{tabular}
		}
\caption{{\bf Frame-level and Video-level comparison on FF\Plus\Plus.}  Multiple metrics reported for medium compression (c23) and high compression (c40) on FF\Plus\Plus~comparing our method with XceptionNet~\cite{rossler2019faceforensics++v3} and DSP-FWA \cite{Li_2019_CVPR_Workshops}. Results are reported on four manipulations.}
\label{tab:ffpp_allmanip}
\end{table}

\minisection{Ablation study, c40} \cref{table:ablation_ffpp_c40} shows the ablation study for different building blocks of the proposed pipeline, along with the proposed loss function. These results are reported without the LSTM thereby evaluating only the backbone without stratified sampling. Given that $D$=$1024$, the two radii are set $r^{\{-,+\}}\doteq\{\on{2.5},\on{97.5}\}$. We report metrics such as TAR and tAUC at a given FAR value. The cut-off FAR points considered are 1\% and 10\%.
At the top, we report results from~\cite{sabir2019recurrent} re-implementing the method without sequential modeling, thereby using just the DenseNet encoder. All methods evaluated in this table consider only the $\net_{\text{RGB}}$ stream and are trained with a global learning rate of $\on{1e-04}$.
Although~\cite{sabir2019recurrent} reaches compelling result at tAUC$_{10\%}$, the performance degrades at lower FAR. Better performance is obtained by combining minor improvements such as training the network with the optimization mentioned in~\cref{sec:arch} that assigns a different updating rate per layer (\emph{+ft}) and dropout (\emph{+drop}) and by using our new loss function (\emph{+loss}) proposed in \cref{sec:loss}. The gain at low false alarm rate is substantial compared to~\cite{sabir2019recurrent} that was trained with binary cross-entropy. In particular, the loss manages to push tAUC$_{1\%}$ up from 57\% to 76\% by imposing a large margin---see  \cref{fig:loss_distance}---while the regular cross-entropy overfits quickly.

\minisection{Ablation study, c23} Given the best results obtained in the previous experiment (i.e., \emph{+ft,+drop,+loss}), we use this configuration as a new baseline to perform other ablations using the FF\Plus\Plus~part with medium compression (c23). \cref{table:ablation_ffpp_c23} reports experiments showing the difference between a single branch and the two-branch structure.

\minisection{Ablation study using LSTM, c40} \cref{table:ablation_ffpp_c40_lstm} shows the ablation experiments when testing the recurrent model. Since adding a recurrent modeling is a drastic change, we verified again that our optimization strategy with different updating rates per layer holds in this case as well. The first two rows in the table support this hypothesis.
We further investigate how to fuse the bi-directional outputs from LSTM and optimize its hidden nodes. Our best result is obtained using hidden node size of 128 and concatenating the two bi-directional outputs. Furthermore, when fusing the two branches \( \net{\text{{\tiny RGB}}}  \circ ~\net{\text{{\tiny LoG}}} \) having convolutional filters divided in two groups is beneficial to the performance.

\minisection{Results} \cref{tab:ffpp_allmanip} shows a thorough comparison on FF\Plus\Plus~\cite{rossler2019faceforensics++v3} training and testing with four manipulations types (Deepfakes, FaceSwap, Face2Face, and NeuralTextures) along with the natural faces. Following~\cite{rossler2019faceforensics++v3}, we trained a model for \texttt{c23} and another for \texttt{c40}.
The table offers multiple evaluations metrics such as AUC, pAUC$_{10\%}$, tAUC$_{10\%}$ and TAR$_{10\%}$. In general, our approach has superior performance compared to Xception. In particular, we improved almost all frame-level performance for the medium compression case (c23), pushing the video-level AUC from 92\% to 99\%. The result is consistent for the other compression level but in general results are lower due to the low image quality; nevertheless our system improves video-level AUC from 86\% to 91\% along with other low false alarm video-level metrics. The table also reports the result of a self-supervised method DSP-FWA~\cite{Li_2019_CVPR_Workshops}.
\cref{tab:ffpp_allmanip_acc} further shows the binary classification accuracies for several state-of-the-art face manipulation detection methods computed on FF\Plus\Plus~\cite{rossler2019faceforensics++v3}. Our approach scores the highest accuracies across manipulations for all the compression levels when trained on the four manipulations. It should be noted that a classifier exploiting the class imbalance here can get an accuracy of 80\% by simply predicting all samples as fakes given that we have 140 real and 560 fake videos or similar balance at the frame level.
\subsection{Celeb-DF}\label{sec:celeb_df}

\begin{table*}[tb]
\centering
\subfloat[]{
	\resizebox{0.65\linewidth}{!}{
		\begin{tabular}{lcccc@{~~~~}cccc@{~~~~}}\toprule
			&  \multicolumn{4}{c}{Frame}  & \multicolumn{4}{c}{Video} \\
			&  \multicolumn{4}{c}{Level}  & \multicolumn{4}{c}{Level}  \\
			Methods & AUC & pAUC$_{10\%}$ & tAUC$_{10\%}$ & TAR$_{10\%}$ & 
			AUC & pAUC$_{10\%}$ & tAUC$_{10\%}$ & TAR$_{10\%}$  \\
 			\cmidrule(l){1-1}
 			Xception-c40~\cite{rossler2019faceforensics++v3} & 65.86  & 54.49 & 12.23 & 22.97 & 69.70 & 57.18 & 16.85 & 34.70 \\
 			DSP-FWA \cite{Li_2019_CVPR_Workshops}             & 64.13  &  52.87 & 10.18 & 19.67 &  69.30 & 51.40 &  17.20 & 32.02 \\
 			Xception-c23~\cite{rossler2019faceforensics++v3} & 66.65 & 53.05 & 10.21 & 19.83 & 73.04 & 52.77 & 9.45 & 18.82 \\
 			\cmidrule(r){1-9} 
 			Ours  & \tbf{73.41} & \tbf{57.42} & \tbf{18.18} &  \tbf{32.22} & \tbf{76.65} & \tbf{58.70} & \tbf{19.73} & \tbf{39.70} \\
			\cmidrule(l){1-9}
		\end{tabular}
		}
    \label{table:celebdf_v2_low_far}
}
\subfloat[]{
    \resizebox{.25\linewidth}{!}{
      \begin{tabular}{l@{~~} c@{~~}c@{~~}}
        \toprule
        Method 
        & FF\Plus\Plus ~\cite{rossler2019faceforensics++v3}
        & Celeb-DF~\cite{li2019celeb_v2}\\
        \cmidrule(r){1-3}
        Two-stream \cite{han2017two}   
        & 70.1 & 53.8  \\
        Meso4 \cite{afchar2018mesonet}
        & 84.7 & 54.8   \\
        MesoInception4 
        & 83.0 &  53.6  \\
        HeadPose \cite{yang2019exposing_uadf}
        & 47.3 & 54.6  \\
        FWA \cite{Li_2019_CVPR_Workshops}
        & 80.1 &  56.9   \\
        VA-MLP \cite{matern2019exploiting}
        & 66.4 & 55.0  \\
        VA-LogReg 
        & 78.0 &  55.1  \\
        Xception-raw \cite{rossler2019faceforensics++v3} 
        & \tbf{99.7} & 48.2  \\
        Xception-c23 
        & \tbf{99.7} & 65.3  \\
        Xception-c40 
        & 95.5 & 65.5  \\
        Multi-task \cite{nguyen2019multi}
        & 76.3 & 54.3  \\
        Capsule \cite{nguyen2019capsule} 
        & 96.6 & 57.5   \\
        DSP-FWA \cite{Li_2019_CVPR_Workshops}
        & 93.0 & 64.6   \\
        \cmidrule(r){1-3}
         Ours  & 93.18 & \tbf{73.41} \\ 
      \bottomrule
      \end{tabular}
    }
    \label{table:celebdf_v2}
}
\caption{{\bf Cross-dataset evaluation on Celeb-DF.} (a) Frame- and video-level performance yet computed at a very low false alarm rate. Best competing methods on Celeb-DF are reported. Ours obtains a wide margin in all the low false alarm rate metrics (b) still performs well when tested on just deepfake class (93.18 \%) AUC on FF\Plus\Plus. Results for other methods are from~\cite{li2019celeb_v2}.
}
\end{table*}

\minisection{Results} We evaluate how well our model transfers to Celeb-DF given that it is trained on FF\Plus\Plus~with multiple manipulations. We do this with the goal of confirming that we optimized our method for better generalization across datasets, reaching a good balance between bias and variance. \cref{table:celebdf_v2_low_far} shows a state-of-the-art evaluation at the frame- and video-level on the 518 test video of Celeb-DF, comparing it to other recent methods. Like other methods~\cite{rossler2019faceforensics++v3}, we trained the model on FF\Plus\Plus~to discern real faces versus four manipulation types at the \texttt{c40} compression level. \cref{table:celebdf_v2_low_far} reports a clear net improvement over the state-of-the-art, even when compared with recent methods that trained the model with self-supervision thereby, in theory, being less prone to overfitting, such as DSP-FWA~\cite{Li_2019_CVPR_Workshops}.
\cref{table:celebdf_v2} offers instead the classic evaluation performance in terms of AUC comparing our approach to the very recent method for digital face manipulation detection. We obtained higher AUC when compared to all the other methods on Celeb-DF while keeping an high AUC on FF\Plus\Plus~on Deepfakes.

\begin{figure*}[tb]
\centering
\subfloat[Natural Sequences]{
    \includegraphics[trim={0 0 0 0},clip,width=0.4\linewidth]{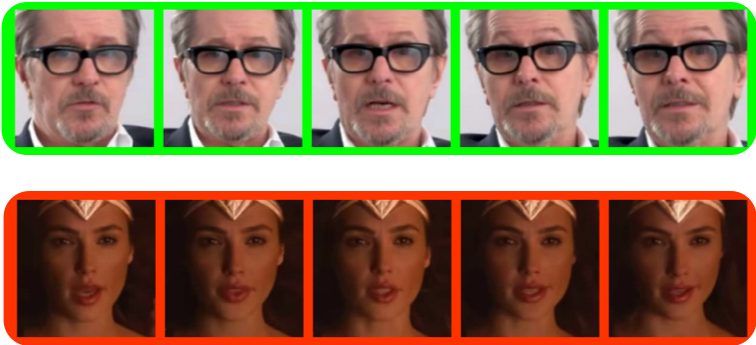}
    \label{fig:real_faces}
}
\subfloat[Manipulated Sequences]{
    \includegraphics[trim={0 0 0 0},clip,width=0.4\linewidth]{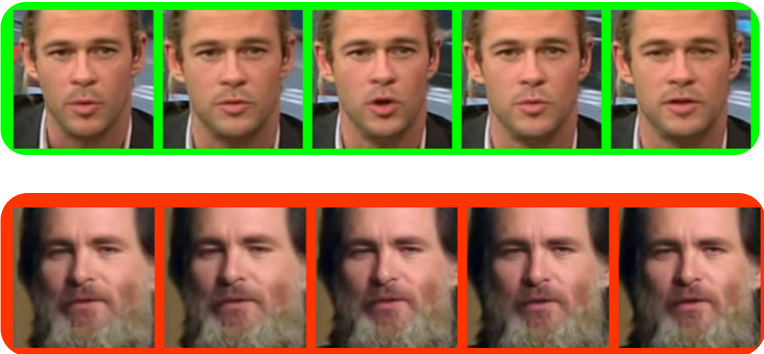}
    \label{fig:fake_faces}
}
\caption{
{\bf Qualitative analysis on Celeb-DF}. The color indicates correct classification (green) or misclassification (red).} 
\label{fig:qualitative}
\end{figure*}

\minisection{Qualitative analysis} \cref{fig:qualitative} shows a qualitative analysis performed on the challenging Celeb-DF~\cite{li2019celeb_v2}. \cref{fig:real_faces} shows untampered faces. The method correctly classifies a sequence with good quality although we used FF\Plus\Plus~with high compression level (c40) for training. Failure case for the natural faces may be caused by the poor illumination. \cref{fig:fake_faces} shows manipulated faces and the method was able to detect a challenging sequence that could be perceived ``as real''; the other failure may be due to the presence of strong facial hairs which could be absent in training data.

\subsection{The Deepfake Detection Challenge (DFDC) Preview Dataset}\label{sec:dfdc}
We report video-level results on the ``The Deepfake Detection Challenge (DFDC) preview set'' using the evaluation described in~\cite{dolhansky_deepfake_2019}. This dataset contains approximately 5,250 videos of digitally manipulated and bona fide videos. As in ~\cite{dolhansky_deepfake_2019}, we used part of the training for cross validation for the two parameters available in our approach that are the optimal number of sequences and the distance $\norm{\net(\bI)-\bc}_2$. We implemented five-fold cross-validation (20\% of training retained for validation) and selected the best pair of parameters across the folds required to maximize the log-weighted precision, log(wP), with $\alpha$=$100$, maintaining the desired level of recall. This procedure was repeated for different cutoff recalls (R$_{10\%}$, R$_{50\%}$, R$_{90\%}$). Although cross validation procedure aims to optimize the two parameters to keep a desired level of recall, meeting the same level of recall is not guaranteed when evaluating on the test set. This procedure simulates what can happen in real scenarios in which a system can be optimized on a validation set and then simply tested in the wild over millions of unlabeled data. 
For this reason, we report log(wP)@recall on the best validation fold under ``valid'' and the test set with ``test-from-valid'' using the parameters from validation. Alternatively, we also searched for the best log(wP) to exactly match the recall value on the test set and report those values under ``test''. 
Except for the above parameter selection, our method has not been re-trained on DFDC preview.

\minisection{Results} \cref{tab:dfdc} shows the evaluation results at the video level. Considering our results under ``test,'' our method has slightly worse precision than XceptionNet~\cite{rossler2019faceforensics++v3} at R$_{10\%}$. However, if we optimize for high recall (R$_{90\%}$), we obtain a substantial boost in the log(wP), increasing log(wP) from -4.041 to -3.548. Moreover, we notice the following if we evaluate with the best hyper-parameters selected on the validation set our method maintains log(wP) better than other methods (-3.721) with a good recall of 0.943.
 
\begin{table}
\centering
\subfloat[]{
	\resizebox{.35\linewidth}{!}{
		\begin{tabular}{lc@{~~}c@{~~}}\toprule
			Methods &  HQ (\texttt{c23})  & LQ (\texttt{c40})\\
			\cmidrule(l){1-1}
			\cite{rossler2019faceforensics++v3}~XceptionNet (Full Image) & 74.78 & 70.52\\
			\cite{fridrich2012rich}~Steg. Features + SVM~  &70.97&55.98   \\
			\cite{Cozzolino17}~Cozzolino~\emph{et al.}~  &	78.45&58.69   \\
			\cite{Bayar16}~Bayar and Stamm     &	82.97&	66.84 \\
		    \cite{Rahmouni2017}~Rahmouni~\emph{et al.}~   &	79.08&	61.18\\
			\cite{afchar2018mesonet}~MesoNet &	83.10	&	70.47	    \\
			\cite{rossler2019faceforensics++v3}~XceptionNet   & {95.73} & {81.00} \\
			\cmidrule(l){1-3}
            Ours  & \tbf{96.43} & \tbf{86.34} \\
			\cmidrule(l){1-3}
		\end{tabular}
		}
	\label{tab:ffpp_allmanip_acc}
}
\subfloat[]{
    \resizebox{.5\linewidth}{!}{
    \begin{tabular}{lc@{\quad}c@{\quad}c@{\quad}c}\toprule
        Method & R$_{10\%}$ & R$_{50\%}$ & R$_{90\%}$\\
        \cmidrule(l){1-1}
        TamperNet~\cite{dolhansky_deepfake_2019} & -2.796@--- & -3.864@--- & -4.041@---\\
        XceptionNet~\cite{rossler2019faceforensics++v3}~(Face)~& \tbf{-1.999@---} & \tbf{-3.012@---} & -4.081@---\\
        XceptionNet~\cite{rossler2019faceforensics++v3}~(Full)~& -3.293@--- & -3.835@--- & -4.081@---\\
        \cmidrule(l){1-4}
        Ours (test) & -2.564@0.100 & -3.152@0.501 & \tbf{-3.548@0.901} \\
        \cmidrule(l){1-4}
        Ours (valid) & -2.311@0.090 & -2.481@0.523 & -2.678@0.918  \\
        Ours (test-from-valid) & -3.386@0.042 & -3.433@0.440 &  -3.721@0.943\\
        \cmidrule(l){1-4}
    \end{tabular}
    \label{tab:dfdc}
    }
}
    \caption{{\bf FF\Plus\Plus~Accuracies and DFDC Preview Dataset.} (a) Comparison of accuracies on FF\Plus\Plus~(b) Video-level $\log(\mathrm{wP})$ for various recall rates.
    }
\label{tab:dfdc_preview_ffpp_acc}
\end{table}

\vspace{-35pt}
\section{Conclusions and Future Work}\label{sec:conclusions}
We presented a method for video-based \deepfake detection that uses a recurrent model to process sequences of aligned faces using a two-branch backbone with a loss function to isolate manipulated face sequences. We have shown results that outperform or are on par with state-of-the-art. However, for practical, web-scale applications, there is significant room for improvement at low false alarm rates. In the short term, we plan to measure the impact of data augmentation and the usage of additional external natural faces~\cite{nagrani2020voxceleb}. In the long term, we plan to augment our model with an explainability mechanism that does not need any pixel-wise supervision for face manipulations.

\minisection{Acknowledgment} This work is based on research sponsored by the Defense Advanced Research Projects Agency under agreement number FA8750-16-2-0204. The U.S. Government is authorized to reproduce and distribute reprints for governmental purposes notwithstanding any copyright notation thereon. The views and conclusions contained herein are those of the authors and should not be interpreted as necessarily representing the official policies or endorsements, either expressed or implied, of the Defense Advanced Research Projects Agency or the U.S. Government. The authors would like to thank E. Sabir and A. Jaiswal for the useful discussions and the anonymous reviewers.

\bibliographystyle{splncs04}

\end{document}